\documentclass[10pt,twocolumn,letterpaper]{article}
\pdfoutput=1
\usepackage[pagenumbers]{wacv} 
\usepackage{comment}
\usepackage{algorithm,algorithmic}
\usepackage{soul}
\usepackage{pbox}
\usepackage{graphicx}
\definecolor{wacvblue}{rgb}{0.21,0.49,0.74}
\usepackage[pagebackref,breaklinks,colorlinks,allcolors=wacvblue]{hyperref}

\def\wacvPaperID{1177} 
\def\confName{WACV}
\def\confYear{2026}

\title{Kinematic-Based Assessment of Surgical Actions in Microanastomosis}

\author{Yan Meng, Daniel Donoho\\
Children's National Hospital\\
DC 20010, USA \\
{\tt\small ymeng@childrensnational.org}
\and
 Marcelle Altshuler, Omar Arnaout\\
Brigham and Women’s Hospital, Harvard Medical School\\
Boston, MA 02115, USA\\
}

\begin{document}
\maketitle
\begin{abstract}
Proficiency in microanastomosis is a critical surgical skill in neurosurgery, where the ability to precisely manipulate fine instruments is crucial to successful outcomes.  These procedures require sustained attention, coordinated hand movements, and highly refined motor skills, underscoring the need for objective and systematic methods to evaluate and enhance microsurgical training. Conventional assessment approaches typically rely on expert raters supervising the procedures or reviewing surgical videos, which is an inherently subjective process prone to inter-rater variability, inconsistency, and significant time investment. These limitations highlight the necessity for automated and scalable solutions. To address this challenge, we introduce a novel AI-driven framework for automated action segmentation and performance assessment in microanastomosis procedures, designed to operate efficiently on edge computing platforms. The proposed system comprises three main components: (1) an object tip tracking and localization module based on YOLO and DeepSORT; (2) an action segmentation module leveraging self-similarity matrix for action boundary detection and unsupervised clustering; and (3) a supervised classification module designed to evaluate surgical gesture proficiency. Experimental validation on a dataset of 58 expert-rated microanastomosis videos demonstrates the effectiveness of our approach, achieving a frame-level action segmentation accuracy of 92.4\% and an overall skill classification accuracy of 85.5\% in replicating expert evaluations. These findings demonstrate the potential of the proposed method to provide objective, real-time feedback in microsurgical education, thereby enabling more standardized, data-driven training protocols and advancing competency assessment in high-stakes surgical environments.
\end{abstract}

\section{Introduction}
\label{sec:intro}

The automated understanding of video content is a long-standing challenge in computer vision and artificial intelligence, with broad applications in domains such as robotics, healthcare, and skill assessment. In high-stakes environments like surgery, the ability to interpret fine-grained actions has direct implications for education, accreditation, and ultimately, patient safety. Among various surgical domains, microanastomosis procedures exemplify the highest levels of technical difficulty, involving the manipulation of ultra-fine instruments under high magnification to suture vessels smaller than 1 mm in diameter. These tasks require precise hand-eye coordination, steady instrument handling, and moment-to-moment motor planning, making them an ideal testbed for AI-driven video understanding.

Traditional methods for evaluating surgical performance rely heavily on expert raters who manually annotate surgical videos using scoring rubrics such as Objective Structured Assessment of Technical Skill (OSATS) \cite{martin1997objective}, Global Rating Scales (GRS) \cite{regehr1998comparing}, and Northwestern Objective Microanastomosis Assessment Tool (NOMAT) \cite{aoun2015pilot}. While informative, such assessments are labor-intensive, subjective, and lack temporal resolution. More recently, there has been growing interest in using machine learning to automate surgical skill assessment \cite{zia2015automated, funke2019video, lavanchy2021automation, meng2023automatic}, often by modeling instrument trajectories, motion dynamics, or gesture sequences. However, many existing approaches focus on coarse-grained scoring, lack interpretability, or fail to generalize across procedures.

In this paper, we focus on the problem of action segmentation and skill assessment in microanastomosis videos, where the goal is to segment surgical procedures into discrete action units and assess performance based on instrument handling behavior. Our approach is inspired by prior work on self-similarity matrices in time-sequence analysis \cite{funke2019video}, which capture temporal changes in action patterns. We integrate this idea with modern deep learning-based object detection and tracking, enabling precise, interpretable segmentation without requiring large-scale labeled datasets for temporal boundaries.

To this end, we propose a novel AI framework that performs temporal segmentation and gesture-level assessment from surgical videos in a fully automated manner. Our contributions are as follows:
\begin{itemize}
	\item Instrument tip tracking based on deep learning algorithms. We leverage You Only Look Once(YOLO) and Deep Simple Online and Realtime Tracking (DeepSORT) models to detect and track instruments, and a customized tip localization algorithm to extract accurate kinematics.
	
	\item Unsupervised action boundary detection using self-similarity matrix. We introduce a novelty function with gaussian-kernel to detect transitions in temporal dynamics, allowing segmentation of continuous motion into pre-defined actions with minimal supervision.
	
	\item Supervised action-level skill classification. We train a classifier with expert-labeled scores to assess the quality of surgical action, enabling interpretable feedback.
	
	\item An end-to-end, modular framework for AI-driven surgical video analysis, designed to operate in real time on edge devices for automated surgical skill assessment. The lightweight system enables medical trainees to perform self-evaluations on their own device. Our method provides a generalizable pipeline for surgical action segmentation and assessment, combining interpretability, precision, and scalability.
\end{itemize}

\section{Related Work}
\label{sec:related_work}

\paragraph{Surgical skill assessment}
The automated evaluation of surgical skill has received growing attention over the past decade due to the increasing demand for standardized, data-driven surgical training. Traditional frameworks such as the OSATS, GRS, and NOMAT provide expert-based evaluations. However, human rating are subjective, time-intensive, and difficult to scale. To address these limitations, early automated assessment efforts focused on analyzing kinematic data from robotic systems such as the Da Vinci Surgical System \cite{ahmidi2017dataset,zia2015automated,ghasemloonia2017surgical}. While effective, these approaches are constrained by hardware availability and limited generalizability to non-robotic procedures.

\paragraph{Instrument detection and tracking}
Tracking surgical instruments is a foundational step in many skill assessment systems. Traditional tracking algorithms relied on handcrafted features \cite{du2016combined}, whereas modern approaches use object detection frameworks based on convolutional neural network, such as CNN, R-CNN, YOLO, and Single Shot Multibox Detector(SSD) \cite{ren2015faster, qiu2019real,redmon2016you,liu2016ssd,fathollahi2022video}. These methods perform frame-level detection and are prone to intermittent failures across time. To address this limitation, multi-object tracking algorithms such as DeepSORT are employed to ensuring temporal continuity and robustness in detection \cite{wojke2017simple}. 

\paragraph{Video-based surgical activity recognition}
Action segmentation is a fundamental task in video understanding that involves partitioning a continuous video sequence into discrete, semantically meaningful units corresponding to specific actions or gestures. In the context of surgical video analysis, this task enables automatic parsing of procedural steps, facilitating downstream tasks such as skill assessment, workflow analysis, and training. 

Traditional approaches have relied on CNNs or 3D CNNs to extract spatiotemporal features from video frames, often combined with recurrent architectures such as LSTMs or GRUs to capture temporal dependencies across frames \cite{donahue2015long,tran2015learning}. These models have been effectively applied to published annotated datasets such as JIGSAWS \cite{gao2014jhu}, Cholec80 \cite{twinanda2016endonet}, and EndoVis Challenge datasets \cite{allan20192017}, primarily for coarse level phase recognition. More recent advancements include Temporal Convolutional Networks (TCNs) \cite{lea2017temporal}, and their multi-stage extension MS-TCN \cite{farha2019ms}, which improve temporal modeling through hierarchical refinement of action boundaries. However, studies have shown that MS-TCN models achieve moderate performance in multiple classes prediction and exhibit limited cross-dataset generalizability, especially when applied to complex or heterogeneous surgical procedures characterized by high procedural variability. \cite{czempiel2020tecno}.

Transformer-based models have demonstrated strong performance in a range of video understanding benchmarks, owing to their capacity for global attention-based temporal reasoning \cite{bertasius2021space,mazzia2022action,zhang2022actionformer, yang2024surgformer}. Despite their high accuracy, these models are computationally demanding, often requiring significant GPU resources and long inference times. Such requirements hinder their deployment in edge environments, including portable surgical training simulators or real-time intraoperative assistance systems, where lightweight and efficient models are critical \cite{wang2022efficient}. Moreover, transformer architectures typically rely on dense temporal supervision and large-scale datasets, which are often unavailable or difficult to get quality label in medical domains.

While the aforementioned methods have demonstrated promising results, their dependence on densely annotated, frame-wise labels and substantial computational resources presents significant practical limitations. Manual annotation at fine temporal granularity is not only labor-intensive and time-consuming but also prone to ambiguity and inter-observer variability, which can introduce noise into the training process. Moreover, their generalizability across surgical domains and procedures is often limited due to overfitting to specific medical dataset distributions.

\begin{figure*}[t]
	\centering
	\includegraphics[width=0.9\linewidth]{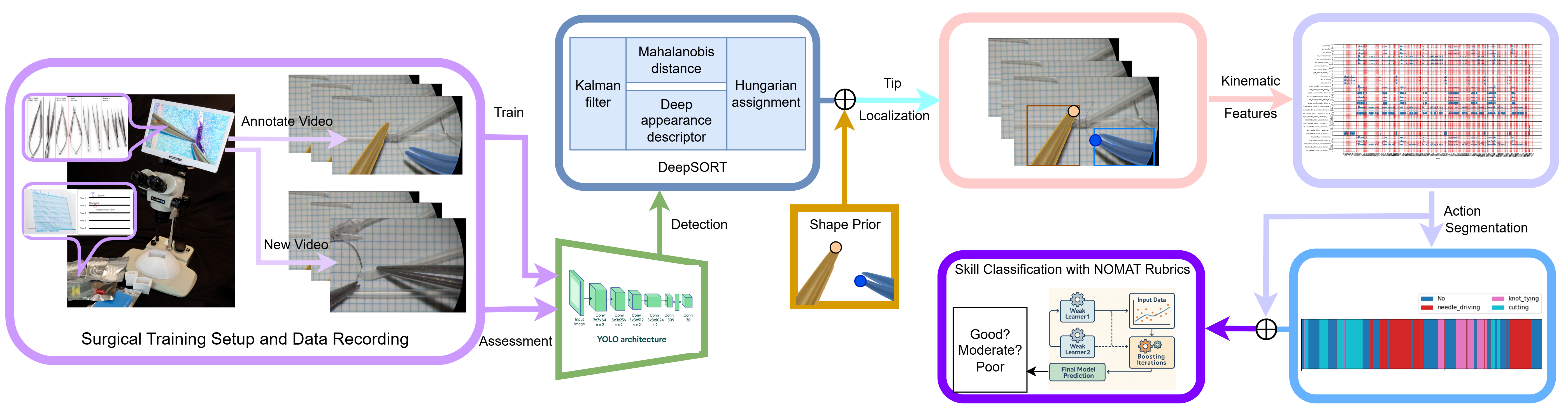}	
	\caption{Microanastomosis skill assessment framework overview.}
	\label{fig:system}
\end{figure*}

\section{Method}
\label{sec:method}
We propose a modular, end-to-end framework for automated action segmentation and surgical skill assessment in microanastomosis videos as show in \Cref{fig:system}. The system consists of three primary components: (1) surgical instrument tip detection and tracking module adapted from YOLO and DeepSORT; (2) unsupervised action segmentation using a novelty function derived from a self-similarity matrix; and (3) supervised classification for action-level skill evaluation. The pipeline is designed to be data and computational efficient, interpretable, and adaptable to various surgical settings.

\subsection{Instrument Tip Detection and Tracking}
Robust and temporally consistent tracking of surgical instruments is critical for extracting meaningful kinematic features for downstream skill assessment. While YOLOv11 provides accurate object detection, its inherently frame-based design can result in inconsistencies across frames, leading to spatial jitter in bounding box coordinates or intermittent detection failures. These artifacts hinder the continuity required for precise surgical motion analysis.

To address this, we adopt a hybrid detection-tracking strategy that integrates YOLOv11 with an enhanced variant of DeepSORT, a multi-object tracker that associates detections using motion and appearance features. The DeepSORT implementation is augmented with a ResNet-based appearance descriptor \cite{he2016deep}, enabling robust identity preservation across frames despite temporary occlusions or missed detections. However, standard DeepSORT suffers from two limitations in our context: (i) it often generates loose bounding boxes that drift from the actual instrument, and (ii) it frequently reinitializes object identities when YOLO temporarily loses or misclassifies a previously seen object. These issues compromise the spatial and temporal fidelity required for fine-grained motion analysis.

To mitigate these limitations, we propose a detection-prioritized ID reassignment mechanism. Specifically:

\begin{itemize}
	\item When both YOLO detections and DeepSORT predictions are available for a frame, we override the tracked bounding box with the corresponding YOLO detection to ensure tighter spatial localization.
	\item To preserve object identity consistency across frames, we concurrently track both class IDs (e.g., needle drivers, scissors) and object IDs throughout the video sequence. In cases where YOLO misclassifies an instrument with DeepSORT successfully maintains a stable object ID, the correct class label can be recovered retrospectively by aligning with the object's prior classification history. Conversely, if DeepSORT erroneously assigns a new object ID due to occlusion or detection dropouts, whereas the class ID remains unchanged, we reassign the original object ID to maintain temporal continuity.
	\item This dual-key identity tracking strategy mitigates the risk of error propagation due to classification drift and reduces dependency on DeepSORT’s internal memory-based object management. As a result, it improves the stability of long-term tracking, reduces memory overhead, and enhances computational efficiency during inference.
\end{itemize}

we estimate the instrument tip position within each tracking bounding box by evaluating a set of candidate points along the instrument shaft. This is achieved by using a shape descriptor that captures local geometric features. Each candidate point's descriptor is compared to a precomputed reference descriptor of the instrument tip using cosine similarity. The point $\hat{p}$ with the highest similarity is selected as the tip location in \Cref{eq:tip}.

\begin{equation}
	\hat{p} = \arg\max_{i \in \mathbf{B}} \frac{\mathbf{D}_{i} \cdot \mathbf{D}_{\text{ref}}}{\|\mathbf{D}_{i}\| \, \|\mathbf{D}_{\text{ref}}\|},
	\label{eq:tip}
\end{equation}

\noindent where $\mathbf{D}_{i}$ is the shape descriptor at candidate point $i$ within bounding box $\mathbf{B}$, and $\mathbf{D}_{\text{ref}}$ is the reference descriptor. 

The selected tip coordinates are subsequently transformed from local bounding box coordinates to the global image space, resulting in a consistent and temporally coherent representation of instrument motion. This multi-stage process ensures precise localization and reliable tracking of surgical instruments, thereby facilitating accurate feature extraction for downstream tasks.

\subsection{Kinematic Change Point Detection}
To segment continuous surgical procedures into discrete action units, we adopt a self-similarity based approach inspired by structure analysis methods in audio and video \cite{foote2000automatic,rodrigues2022feature}, where change points in the multivariant time sequence is detected with novelty function. 
\paragraph{Kinematic features self-similarity matrix} From the tracked instrument tip positions in a surgical video of length $T$, a kinematic feature vector $\mathbf{X} \in \mathbb{R}^{T \times d}$ is extracted at each video frame, comprising the velocity, acceleration, and jerk of each individual instrument, as well as inter-instrument distance, relative velocity, dot product, and angle. To partition the continuous video stream into discrete surgical actions, we construct a self-similarity matrix (SSM) by computing pairwise cosine similarity between the normalized kinematic feature vectors:

\begin{equation}
	S_{i,j} = \frac{\mathbf{X}_i \cdot \mathbf{X}_j}{\|\mathbf{X}_i\| \, \|\mathbf{X}_j\|}, \quad \text{for } i,j = 1, \dots, T
	\label{eq:ssm}
\end{equation}

To enhance the contrast of block-diagonal structures in the SSM, we apply a nonlinear contrast enhancement to the matrix $ S^{'}=S^{2}$

\paragraph{Change point detection} We compute a novelty function over time by convolving a Gaussian kernel over the main diagonal of the similarity matrix as in \Cref{eq:novelty}.

\begin{equation}
	N(t) = \sum_{i=-h}^{h-1} \sum_{j=-h}^{h-1}G(i,j) \cdot C(i,j) \cdot S^{'}(t+i, t+j)
	\label{eq:novelty}
\end{equation}

\noindent where $G(i,j)$ is a 2D Gaussian function, $C(i,j)$ defines the checkerboard structure with alternating positive and negative quadrants, and $h$ is half of the kernel size.

This design introduces temporal weighting through a Gaussian function while preserving the core concept of the original novelty method. The Gaussian kernel generates a smoother and more stable novelty curve, which enhances the reliability of peak detection and reduces false positives in boundary localization. It suppresses the influence of temporally distant similarity values, and the emphasis on nearby temporal structure allows for more precise alignment with transitions in fine-grained surgical actions.

To identify action boundaries from the novelty function, we apply a peak-picking algorithm based on local maxima detection \cite{gokcesu2021nonparametric}. Given a discrete novelty function $N(t) = \{ N_t\}_{t=1}^{T}$, the goal is extract a set of time indices $\boldsymbol{\tau} = \{\tau_1, \tau_2,..., \tau_k\}$ such that each $\tau_i$ corresponds to a local maximum that satisfies certain prominence $\pi(t)$ and distance criteria $d_{min}$. The prominence $\pi(t)$ of a peak at time $t$ is defined as the vertical distance between the peak height and the highest minimum on either side in \Cref{eq:prominence}.
\begin{equation}
	\pi(t) = N_t - \max\left( \min_{t^{-} < i < t} N_i,\ \min_{t <j <t^{+}} N_j \right)
	\label{eq:prominence}
\end{equation}

\noindent where $t^{-}$ and $t^{+}$ are the boundaries of the peak's neighborhood. A peak is retained only if $\pi(t)$ is greater than a given threshold. A minimum distance $ |\tau_{i+1} -\tau_i |\geq d_{min}$ is imposed between consecutive peaks to prevent detection of densely packed local maxima.

\subsection{Action Clustering}
Following the detection of temporal boundaries using the novelty function, the surgical video is segmented into atomic units. Descriptive features are then extracted from each segment to facilitate unsupervised clustering. This process enables the grouping of similar surgical actions without the need for manual annotation, thereby significantly reducing the time and effort required from expert surgeons.

Given the kinematic feature vector $\mathbf{X} \in \mathbb{R}^{T \times d}$ and the segment boundaries (\ie peaks in the novelty function) $\mathcal{B} =\{b_{0}, b_{1}, ..., b_{N}\}$, where $b_{0} = 0$ and $b_{N} = T$, we define segment $s_{i}$ as interval $\left({b_{i},b_{i+1}} \right] $ for $ i=0,..., N-1$. For each segment $s_{i}$, we compute the feature summary vector as in \Cref{eq:statistic_summary}.

\begin{equation}
	\mathbf{f}_i = \boldsymbol{\mu}_{i} \oplus \boldsymbol{\sigma}_{i} \oplus \boldsymbol{Mask}_{i}
	\label{eq:statistic_summary}
\end{equation}

\noindent where $\oplus$ is a concatenation operation, $\boldsymbol{\mu}_{i}$ and $\boldsymbol{\sigma}_{i}$ is the mean and standard deviation of the segment respectively, $\boldsymbol{Mask}_{i}$ is the indicator of instrument presence in the scene, which is semantically informative to the action classification.

We apply K-means clustering to the feature summary matrix $\mathbf{F}$ with an objective function that minimize within-cluster variance:
\begin{equation}
	\underset{\{C_k\}_{k=1}^{K}}{\arg\min} \sum_{k=1}^{K} \sum_{\mathbf{f}_i \in C_k} \|\mathbf{f}_i - \mathbf{c}_k\|^2,
	\label{eq:kmeans}
\end{equation}

\noindent where $C_k$ denotes the set of feature vectors assigned to cluster $k$, and $\mathbf{c}_k$ is the centroid of cluster $k$.
\newline

\subsection{Microanastomosis Skill Classification}
Each video captures a complete microanastomosis procedure, comprising three vessel cuts followed by eight suture placements. To evaluate surgical performance, we adopt the NOMAT rubric as an evaluation standard \cite{aoun2015pilot}, and emulate expert grading using a supervised machine learning approach, where ground truth labels are provided by board-certified neurosurgeons. For skill classification, we employ a Gradient Boosting Classifier (GBC) \cite{konstantinov2021interpretable}, chosen for its ability to model complex, non-linear decision boundaries while maintaining robustness to overfitting on small to medium-sized datasets. The classifier takes as input a feature vector constructed from each action, including the feature summary in \Cref{eq:statistic_summary}, action label, number of repetitions, and duration of each action. The output is a categorical skill label: Poor, Moderate, or Good, corresponding to the expert-rated performance levels.

The complete workflow of the proposed method is outlined in \Cref{alg:action_segmentation}.

\begin{algorithm}[t]
	\caption{Microanastomosis Skill Assessment}
	\label{alg:action_segmentation}
	\begin{algorithmic}
		\REQUIRE A set of surgical video $\mathcal{V}$
		\vspace{0.2em}
		\STATE  \underline{\textbf{Instrument Tips Detection \& Tracking}}
		\FOR{each frame $t = 1$ to $T$}
		\STATE $B_t \gets \textsc{YOLO}(I_t) $ \hfill // Detect bounding boxes 
		\STATE $ \mathcal{T}_t \gets \textsc{DeepSort}(B_t, \mathcal{T}_{t-1}) $ \hfill // Assign consistent IDs
		\STATE $ \mathbf{p}_t \gets \textsc{MatchTip}(B_t, \mathcal{T}_t,\Theta_{\text{descriptor}}) $ \hfill // Locate tip
		\ENDFOR
		\STATE Construct and normalize features $\mathbf{X } \in \mathbb{R}^{T \times d}$ from tips
		\vspace{0.5em}
		
		\STATE \underline{\textbf{Motion Change Point Detection}}
		
		\STATE $S \gets \textsc{SSM}(\mathbf{X })$	\hfill //Compute self-similarity matrix
		\STATE $S^{'} = S^2$		\hfill  // Enhance contrast
		\STATE $N(t) \gets \textsc{Novelty}(S^{'}) $ \hfill  // Compute novelty function
		\STATE $\mathcal{B} \gets \textsc{Peaks}(N(t))$  	\hfill // Find motion boundaries
		\vspace{0.5em}
		
		\STATE \underline{\textbf{Action Clustering}}
		\FOR{each segment $s_{0}$ to $s_{N-1}$}
		\STATE $\mathbf{f}_i \gets \textsc{Features} (s_{i}) $
		\ENDFOR 			
		\STATE $C_k \gets \textsc{K-means}(\{\mathbf{f}_i\})$ \hfill //Action classification
		\vspace{0.5em}
		
		\STATE \underline{\textbf{Skill Classification}}
		\STATE $Score \gets \textsc{GBC}(\{\mathbf{f}_i,N_{repeat}, Duration_i, Action\})$
	\end{algorithmic}
\end{algorithm}
\section{Experiments}
\label{sec:expr}
We conducted a comprehensive experimental study to evaluate the effectiveness of the proposed AI-based surgical skill assessment framework. All data collection procedures and study protocols were reviewed and approved by the Institutional Review Board of our collaborating institution, ensuring compliance with ethical standards for research involving human subjects.

\subsection{Data Curation}

Microanastomosis procedures were conducted within a controlled simulation environment using standardized microsurgical training kits. Each setup included a Meiji Techno EMZ-250TR trinocular zoom stereomicroscope, equipped with a high-definition camera and external display monitor to facilitate data recording and observational analysis. Vascular tissues were simulated by 1.0mm $\times$ 0.8mm microvascular anastomosis practice cards (Pocket Suture, USA), designed to replicate the dimensions and mechanical properties of small-caliber vessels. A uniform set of microsurgical instruments was supplied to all participants to ensure procedural consistency, including one straight needle driver, one curved needle driver, and a pair of both straight and curved microsurgical scissors. The complete simulation setting is illustrated in \Cref{fig:system}.

A total of nine medical practitioners (8 male, 1 female; average age of 30.5 years) participated in this study, representing a broad spectrum of microsurgical experience, ranging from novice trainees to expert neurosurgeons. All participants were predominantly right-handed, with the exception of one individual who identified as ambidextrous. Each procedure followed a consistent protocol, comprising three vessel transection, followed by a sequence of eight suture placements, enabling consistent analysis across sessions. Each participant performed between five and ten microanastomosis procedures under standardized laboratory conditions, resulting in a total of 63 video recordings. Of these, 58 recordings captured the entire procedure from start to finish, with an average duration of approximately 26 minutes, while five recordings were incomplete due to technical interruptions or early termination. 

Two board-certified neurosurgeons independently evaluated instrument handling proficiency in each video using the NOMAT rubric and subsequently reached consensus on all assessments. Performance was rated on a five-point Likert scale, with higher scores reflecting greater levels of technical proficiency.

\subsection{Instrument Detection Model Training}
\begin{figure}[t!]
	\centering
	\includegraphics[width=0.95\linewidth]{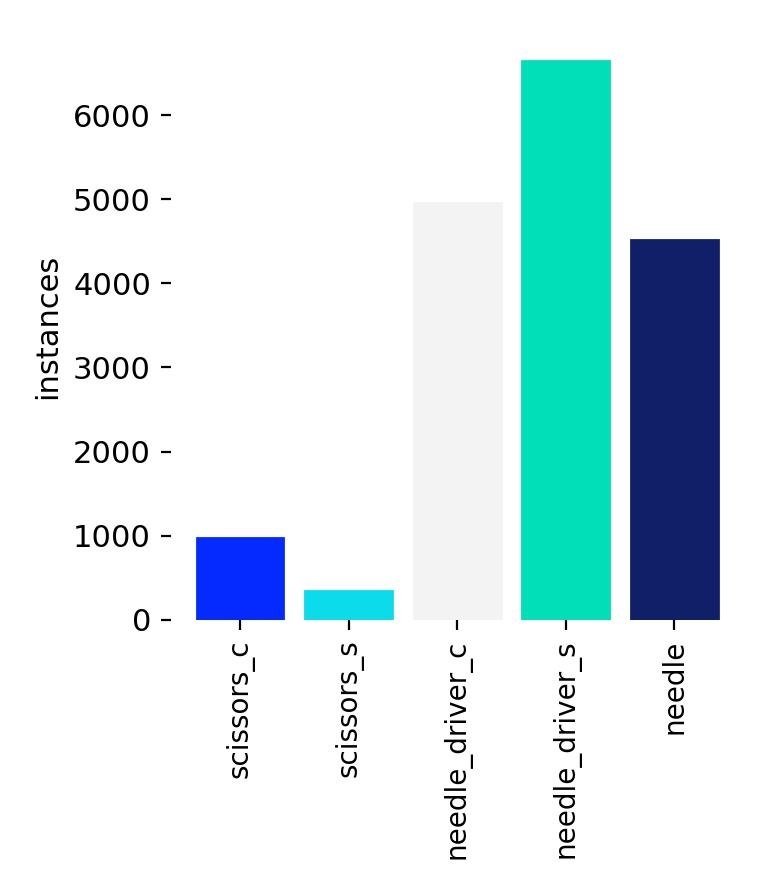}
	\caption{Instrument detection class distribution in the dataset}
	\label{fig:yolo_classes}
\end{figure}

To train the object detection model, a subset of four videos was selected and downsampled, resulting in a total of 47,344 frames. This image dataset was partitioned into training, validation, and test sets using an 80\%:10\%:10\% split. Ground truth annotations for surgical instruments were labelled semi-automatically using the Encord platform, in combination with the Segment Anything Model (SAM)\cite{kirillov2023segment}. The class-wise distribution of annotated instances is illustrated in \Cref{fig:yolo_classes}.

\begin{figure}[t!]
	\centering
	\includegraphics[width=0.9\linewidth]{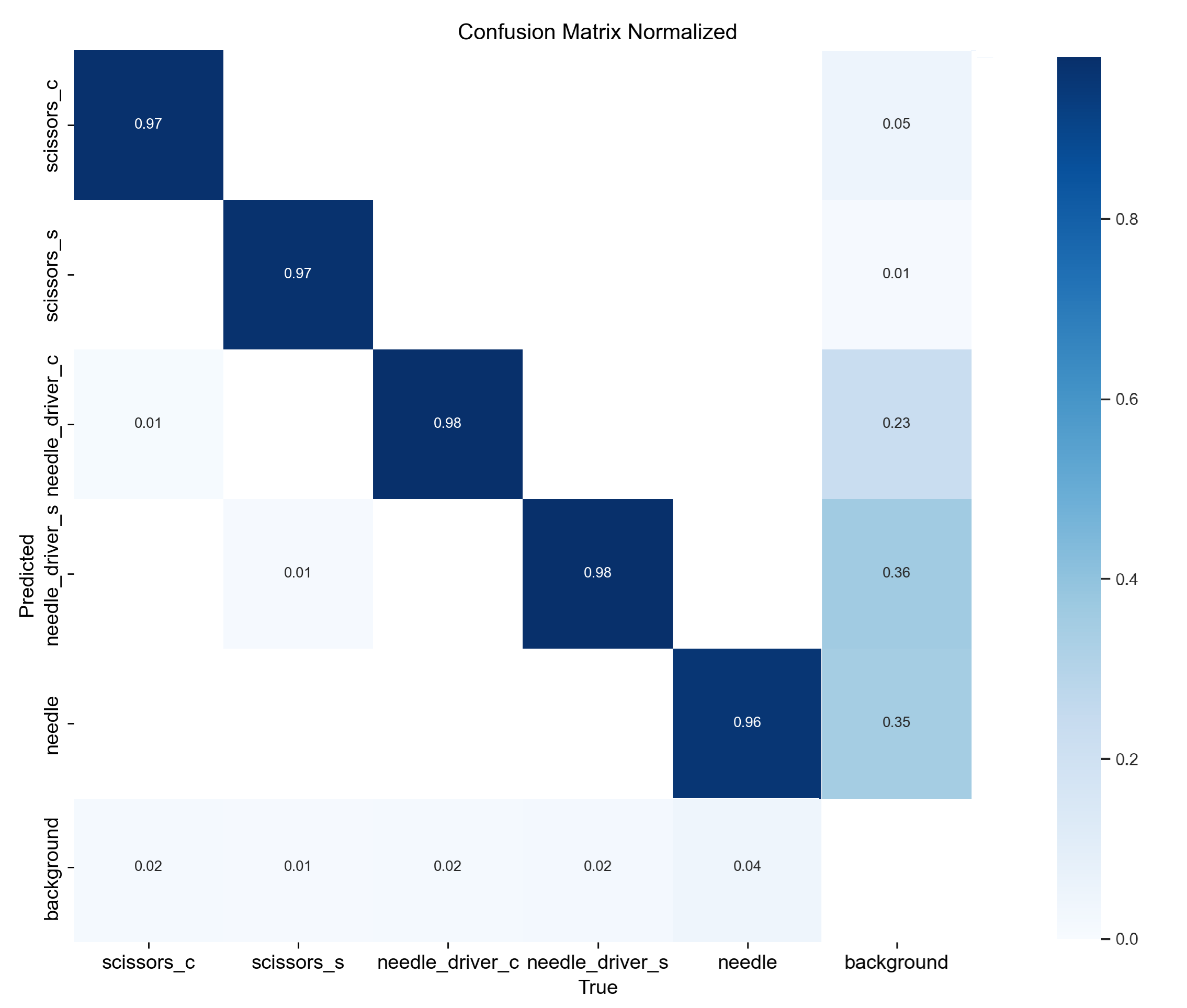}
	\caption{Normalized class confusion matrix.}
	\label{fig:confusion}
\end{figure}

\begin{figure*}[t]
	\centering
	\begin{subfigure}{0.3\textwidth}
		\includegraphics[width=\textwidth]{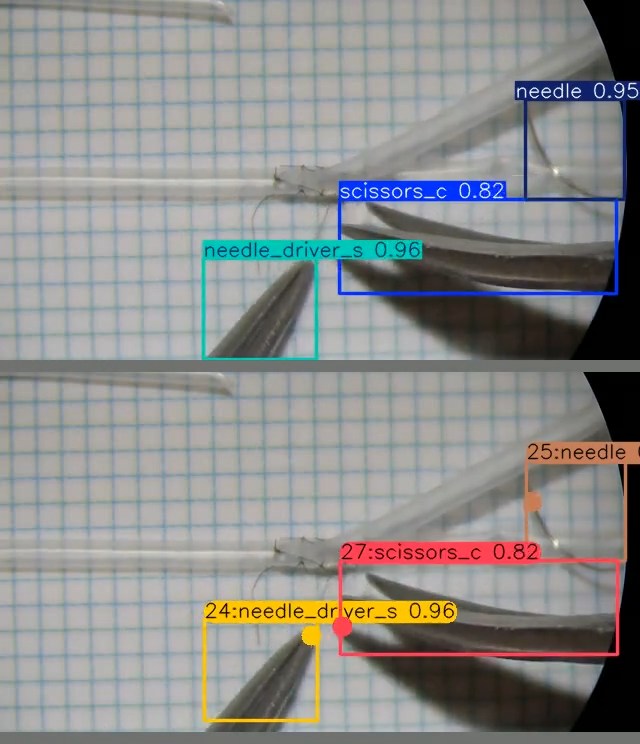}
		\caption{Participant 0824 frame 4191.}
		\label{fig:frame1}
	\end{subfigure}
	\hfill
	\begin{subfigure}{0.3\textwidth}
		\includegraphics[width=\textwidth]{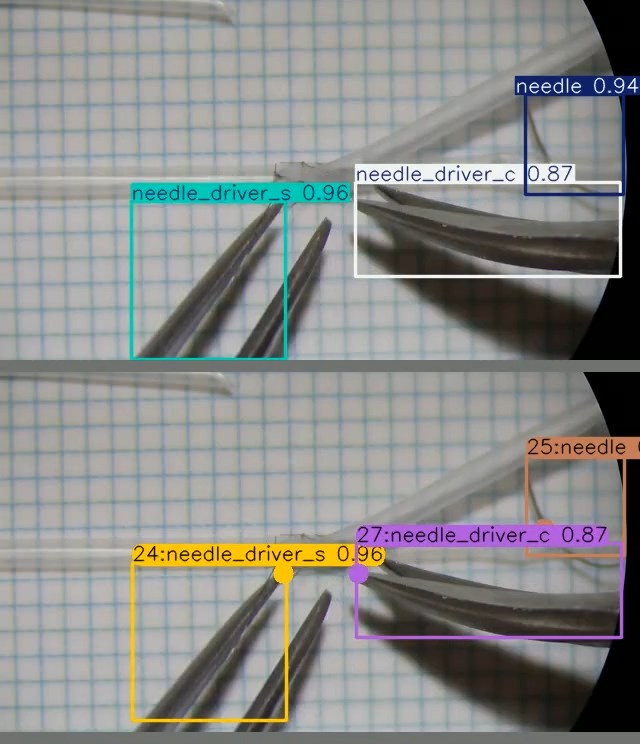}
		\caption{Participant 0824 frame 4192.}
		\label{fig:frame2}
	\end{subfigure}
	\hfill
	\begin{subfigure}{0.3\textwidth}
		\includegraphics[width=\textwidth]{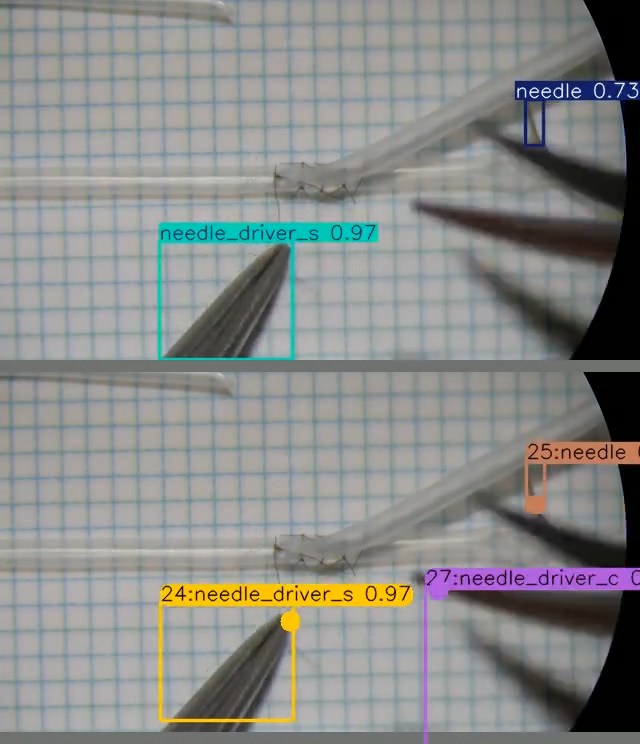}
		\caption{Participant 0824 frame 4198.}
		\label{fig:frame3}
	\end{subfigure}
	\caption{Comparison of the proposed tip tracking method with YOLOv11, the top row is the YOLOv11 object detection results, the bottom row is the proposed instrument tracking and tip localization results.\label{fig:yolo_sort}}
\end{figure*}

The object detection model was trained using the YOLOv11 architecture for 100 epochs on a consumer computer equipped with a 12th Gen Intel® Core™ i9 processor @2.50 GHz and a NVIDIA GeForce RTX 3080 Ti laptop GPU. Training was conducted using stochastic gradient descent with early stopping monitored via validation loss to avoid overfitting. The loss function and associated performance metrics, including precision, recall, and mean Average Precision, exhibited stable convergence after approximately 50 epochs. 

To evaluate class-specific detection performance, we present the normalized confusion matrix in \Cref{fig:confusion}, which highlights the model’s ability to distinguish among different instrument classes with minimal inter-class confusion. Additionally, the aggregate instrument detection performance metrics across all classes is summarized in \Cref{tab:yolo}, reflecting consistent high-accuracy detection performance.

\begin{table*}[t!]
	\centering
	\begin{tabular}{@{}lcccccc@{}}
		\toprule
		Class    &No. of Images     & Total Instances      & Precision      & Recall  & mAP50  & mAP50–95	\\
		\midrule
		all      & 4174	    		& 9417    	&  0.969   & 0.966  	& 0.989    & 0.958 \\
		scissors\_c     & 398      & 398    	& 0.981  	& 0.96		&0.992		& 0.976  \\
		scissors\_s     & 70     	& 70   		& 0.921  	& 1			&0.995		& 0.979 	\\
		needle\_driver\_c     & 2861  & 2861	& 0.989  	& 0.975		&0.993		& 0.973	\\
		needle\_driver\_s     & 3669  & 3779	& 0.986  	& 0.975		&0.992		& 0.965 \\
		needle     & 2309  & 2309	& 0.971  	& 0.918		&0.972		& 0.872 	\\
		\bottomrule
	\end{tabular}
	\caption{Instrument detection performance metrics.}
	\label{tab:yolo}
\end{table*}

\subsection{Instrument Tip Tracking}
\label{sec:itt}
The performance of the improved DeepSORT tracking algorithm is evaluated in terms of its ability to recover missed instrument detections and correct misclassified labels produced by the YOLO detector. We define two complementary evaluation metrics: Recovery Rate (RR) quantifies the percentage of missed detections that were successfully re-identified by the tracking system, and Correction Rate (CR) measures the proportion of incorrectly labeled instruments whose class identities were rectified by the tracker. As summarized in Table \ref{tab:deepsort}, the improved DeepSORT achieved an average recovery rate of 98.7\% and a correction rate of 90.6\%, indicating strong reliability in maintaining temporal consistency and resolving detection errors.

\begin{table}[t]
	\centering
	\begin{tabular}{@{}lcc@{}}
		\toprule
		Class    &Recover Rate     & Correction Rate   	\\
		\midrule
		all      & 0.987	& 0.906	 \\
		scissors\_c     	& 0.992	& 0.864  \\
		scissors\_s     &0.995	& 0.872	\\
		needle\_driver\_c      &0.989	& 0.902	\\
		needle\_driver\_s     &0.994	& 0.895 \\
		needle     & 0.963	& 0.997	\\
		\bottomrule
	\end{tabular}
	\caption{Instrument tracking algorithm performance.}
	\label{tab:deepsort}
\end{table}

Following instrument tracking, the tip localization algorithm is applied to the temporally consistent bounding boxes. \Cref{fig:yolo_sort} illustrates representative frames from a test sequence to demonstrate the effectiveness of the integrated tracking and tip localization pipeline. In \Cref{fig:frame1}, the YOLO detector correctly identifies a curved scissors; however, in the following frame \Cref{fig:frame2}, the same instrument is misclassified as a curved needle driver by YOLO. Despite the misclassification, the tracker maintains a consistent object ID across frames based on temporal context. Furthermore, \Cref{fig:frame3} shows a case where the detector entirely misses the scissors in a given frame. The tracking module not only recovers this instance but also preserves its identity throughout the subsequent frames. In all cases, the tip localization algorithm successfully identifies the tips of instruments, demonstrating high spatial accuracy under both ideal and error-prone detection conditions.

\begin{figure}[t!]
	\centering
	\includegraphics[width=\linewidth]{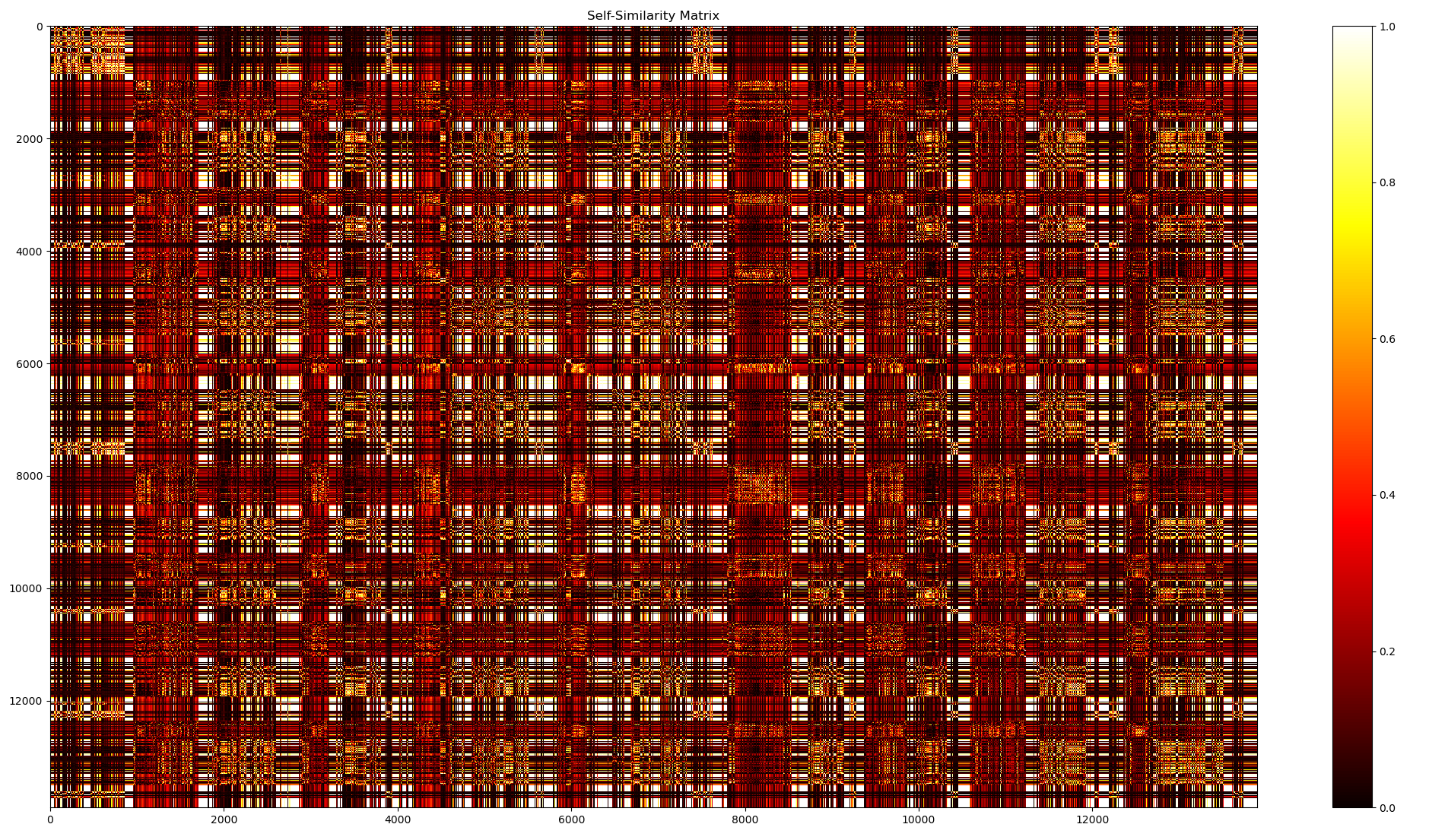}
	\caption{Kinematic feature self-similarity matrix.}
	\label{fig:ssm}
\end{figure}

\begin{figure*}[t!]
	\centering
	\includegraphics[width=\linewidth]{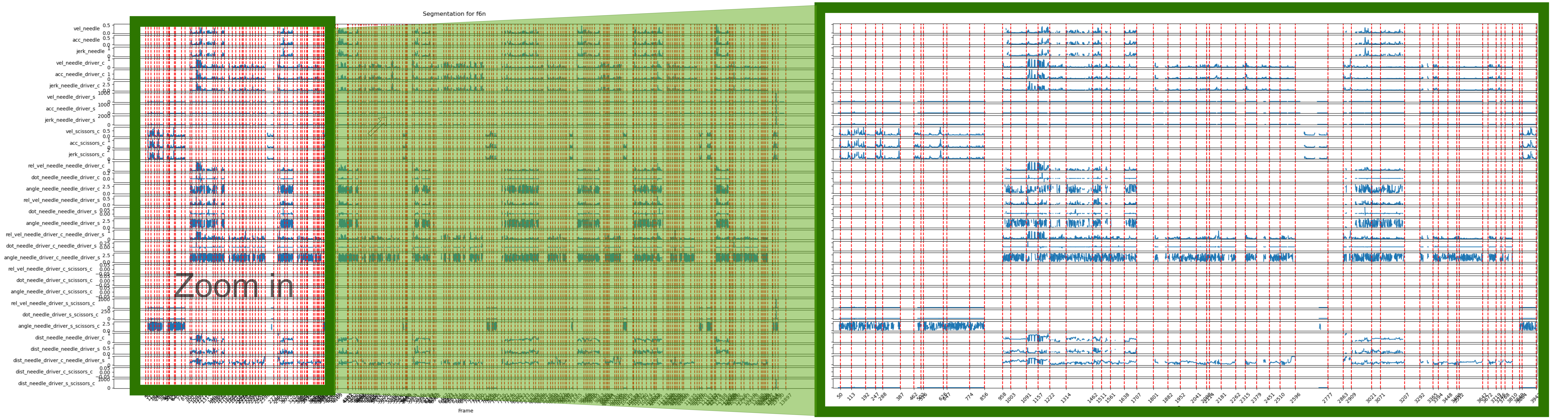}
	\caption{Visualization of candidate action boundaries in the microanastomosis procedure. The left panel presents a series of subplots illustrating the temporal evolution of each kinematic feature across the entire procedure, while the right panel provides a magnified view focusing on the first 40 seconds to highlight finer-grained temporal patterns. }
	\label{fig:change_point}
\end{figure*}

\begin{figure*}[t!]
	\centering
	\includegraphics[width=\linewidth]{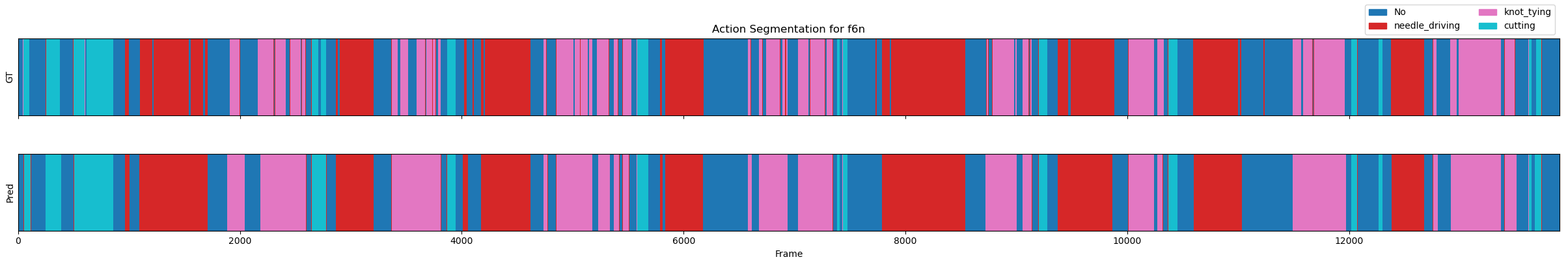}
	\caption{An example of the action segmentation results compared with ground truth label.}
	\label{fig:seg_illustrition}
\end{figure*}

\subsection{Kinematic-based Action Segmentation}

The microanastomosis procedure follows a standardized sequence of microsurgical steps, beginning with vessel preparation and followed by suturing. Specifically, the procedure initiates with the excision of the donor vessel, after which the tip of the donor vessel is incised longitudinally to create an opening. Subsequently, the recipient vessel is also incised to allow for an end-to-side anastomosis, enabling the donor vessel to be sutured in place. The suturing phase consists of eight stitches, each comprising a sequence of surgical actions: needle driving, knot tying, and thread cutting.

For the purpose of action segmentation and classification, we define three primary action classes corresponding to semantically meaningful microsurgical actions: \textbf{Cutting}, \textbf{Needle Driving}, and \textbf{Knot Tying}. In addition, we include a fourth class, labeled as \textbf{No Action} to group frames that have no instrument in the scene, or the instruments moving in the background with no clear actions.

Instrument tip trajectories are extracted as described in \Cref{sec:itt}, enabling computation of low-level kinematic features such as velocity, acceleration, jerk, relative distance, velocity and angles between instruments. These features are then embedded into a self-similarity matrix (SSM), capturing frame-wise temporal relationships based on cosine similarity between feature vectors. An example visualization of the SSM for a microanastomosis video is shown in Figure~\ref{fig:ssm}, highlighting the repeating structure of kinematic features in the time sequence. Peaks in the computed novelty function indicate significant transitions in motion dynamics, which we interpret as candidate boundaries for action segmentation. Following the detection of change points, the K-means algorithm groups temporally similar actions by assigning identical cluster ID to instances of the same action class. To evaluate the performance of the unsupervised clustering algorithm, we manually annotated 20 videos with frame-level action labels. The quantitative results of the segmentation is reported in \Cref{tab:action}, where our method achieves an action segmentation accuracy of 92.41\%. A qualitative comparison between the unsupervised action segmentation and ground truth annotations for an example video is illustrated in \Cref{fig:seg_illustrition}. This result outperforms previously reported performance from state-of-the-art methods such as MS-TCN \cite{czempiel2020tecno} and transformer-based models \cite{yang2024surgformer} on several publicly available datasets. However, due to the unavailability of kinematic information in these benchmark datasets, a direct comparison using our feature set could not be conducted.

\begin{table}[t]
	\centering
	\begin{tabular}{@{}lccccc@{}}
		\toprule
		Test set    &Acc &Prec  & Rec  & F1 	& Jacc \\
		\midrule
		20 videos      & 0.924	& 0.764 & 0.824 &0.790	&0.735	 \\		
		\bottomrule
	\end{tabular}
	\caption{Action segmentation method performance.}
	\label{tab:action}
\end{table}

\begin{table}[t!]
	\centering
	\resizebox{\columnwidth}{!}{
		\begin{tabular}{@{}lcccccc@{}}
			\toprule
			&\multicolumn{3}{c}{Needle Driving}  &\multicolumn{3}{c}{Knot Tying}\\
			\cmidrule(lr){2-4} \cmidrule(lr){5-7}
			Level     &Prec  & Rec  & F1    &Prec  & Rec  & F1 \\
			\midrule
			Poor     	&0.89	& 1.00 	& 0.94 	&0.89	& 1.00 	& 0.94\\
			Moderate	& 0.80	&0.97	& 0.88	 & 0.80	&0.97	& 0.88   \\
			Good		&1.00 	& 0.67 	& 0.80 	&1.00 	& 0.67 	& 0.80 \\	
			\bottomrule
		\end{tabular}
	}
	\caption{Needle driving and knot tying action skill evaluation.}
	\label{tab:nd_kt}
\end{table}

\subsection{Microanastomosis Skill Classification}
We utilize action-level kinematic features, along with the repetition count and temporal duration of each action, to train a supervised classification model based on gradient boosting. A total of 58 complete microanastomosis procedures were used for model development. The dataset was partitioned into an 80\% training set and a 20\% testing set. Within the training set, five-fold cross-validation was employed, where the model was iteratively trained on four folds and validated on the fifth to ensure generalization and robustness. Given the relatively small dataset and the imbalanced distribution of expert-assigned scores, the original rating scale was discretized into three ordinal classes: Poor, Moderate, and Good, using score thresholds of 2.5 and 3.5. This regrouping resulted in class distributions of 32, 15, and 11 samples, respectively, for the needle driving action, and 37, 10, and 11 samples for the knot tying action. The classification results for each action type are summarized in \Cref{tab:nd_kt}. The gradient boosting classifier achieved an accuracy of 84\% on needle driving actions and 88\% on knot tying actions, resulting in an average classification accuracy of 85.5\% across both action types. These findings demonstrate that the proposed system is capable of performing interpretable, action-level skill assessment. 

\section{Conclusion}
\label{sec:conclusion}
In this paper, we present an AI-driven framework for automated action segmentation and skill assessment in microanastomosis procedures, addressing the need for objective, scalable, and fine-grained evaluation in surgical training. Our system integrates deep learning based instrument detection and tracking, unsupervised action segmentation, and supervised classification of surgical skill. By segmenting complete procedures into discrete action-level clips and extracting meaningful kinematic and temporal features, the proposed method emulates expert surgical assessments at a high temporal resolution. Moreover, our method offers grading outputs that align with established NOMAT surgical assessment rubrics, enabling interpretable feedback for trainees. The modularity of the framework allows for future extensions and generalization to other surgical tasks or procedural domains.

{
    \small
    \bibliographystyle{ieeenat_fullname}
    \bibliography{ref}
}

\end{document}